# Dynamic Dimension Wrapping Algorithm: A Novel Approach for Efficient Cross-Dimensional Search in Dynamic Multidimensional Spaces

Dongnan Jin[a], Yali Liu[a, *], Qiuzhi Song[a, b], Xunju Ma[a], Yue Liu[a], Dehao Wu[a]

[a]School of Mechatronical Engineering, Beijing Institute of Technology, Beijing 100081, China
[b]Advanced Technology Research Institute, Beijing Institute of Technology, Jinan 250300, China
* Corresponding author.
E-mail address: dongnanjin@163.com (D. Jin), buaaliuyali@126.com (Y. Liu), qzhsong@bit.edu.cn (Q. Song), xunjuma@163.com (X. Ma), ly2078917188@163.com (Y. Liu), deehao.wu@gmail.com (D. Wu).

**Abstract.** To effectively search for the optimal motion template in dynamic multidimensional space, this paper proposes a novel optimization algorithm, Dynamic Dimension Wrapping (DDW). The algorithm combines Dynamic Time Warping (DTW) and Euclidean distance, and designs a fitness function that adapts to dynamic multidimensional space by establishing a time-data chain mapping across dimensions. This paper also proposes a novel update mechanism—Optimal Dimension Collection (ODC), combined with the search strategy of traditional optimization algorithms, enables DDW to adjust both the dimension values and the number of dimensions of the population individuals simultaneously. In this way, DDW significantly reduces computational complexity and improves search accuracy. Experimental results show that DDW performs excellently in dynamic multidimensional space, outperforming 31 traditional optimization algorithms. This algorithm provides a novel approach to solving dynamic multidimensional optimization problems and demonstrates broad application potential in fields such as motion data analysis.

## 1 Introduction

Human motion analysis has important applications in various disciplines, such as biomechanics, sports science, and rehabilitation engineering. In order to effectively understand the dynamic characteristics of the human body under different movement patterns, studies typically rely on complex human-machine coupling models to process time data sequences related to gait. However, optimization algorithms can simplify this problem into a nonlinear programming (NLP) [1] problem, thereby bypassing the complex human-machine coupling models. Within this framework, the optimization algorithm searches for an optimal time data chain to serve as the optimal motion template, minimizing the distance between this chain and the original gait data. This optimal motion template can represent all the typical dynamic characteristics under a specific movement pattern, thereby providing effective theoretical support for future human motion research.

For NP-hard optimization problems in the real world, Meta-heuristic (MH) optimization algorithms provide strong guidance for the development of optimization strategies, while demonstrating exceptional search capabilities. In previous studies, scholars have proposed many excellent metaheuristic algorithms which have been widely applied in various fields, such as medicine [2], engineering [3], and finance [4,5]. MH can be divided into the following four categories: Swarm intelligence (SI) [6] algorithms simulate the intelligent coordination and social behavior of a swarm or group, such as Particle Swarm Optimization (PSO) [7], Gray Wolf Optimizer (GWO) [8], Moth Flame Optimization (MFO) [9], Aquila Optimizer (AO) [10], Dingo Optimization Algorithm (DOA) [11];Evolutionary algorithms (EA) mimic the evolutionary behavior of natural organisms, such as Genetic Algorithms (GA) [12], Differential Evolution (DE) [13] algorithms, and Evolutionary Strategies (ES) [14];Physics-based (PB) algorithms mimic fundamental laws of physics or physical phenomena, such as Simulated Annealing algorithm (SA) [15], Gravitational Search Algorithm (GSA) [16], Archimedes Optimization Algorithm (AOA) [17], and Water Wave Algorithm (WWO) [18];Human-based (HB) algorithms related to human behavior, such as Harmony Search (HS) [19]. Despite the existence of several excellent optimization algorithms, according to the "No Free Lunch" (NFL) theorem [20], no single optimization method can achieve the best solution for all types of problems [21]. Therefore, the development of new optimization algorithms to address emerging practical problems holds significant research significance and practical value.

The traditional optimization algorithms mentioned above are generally effective at handling problems with a fixed number of dimensions (as shown by "MH" in Figure 1). Their search process operates within a multi-dimensional space of constant dimensionality, aiming to find the optimal variable combination with a fixed number of dimensions. In this process, each individual in the population has the same, fixed number of dimensions. However, in gait data analysis, due to the partial asymmetry of human motion, we observe that the number of dimensions in the multi-dimensional space represented by the gait time series changes dynamically. As illustrated by "DDW" in Figure 1, the number of dimensions in this multi-dimensional space, formed by the raw gait data, is variable, i.e., a dynamic multi-dimensional space. This dynamic space can be decomposed into several subspaces, each with a fixed number of dimensions, where the desired optimal motion template may exist in any of these subspaces. Unless bearing enormous computational costs and conducting exhaustive searches in each subspace, traditional optimization algorithms will inevitably overlook the optimal solutions in other multidimensional spaces and may very likely fail to obtain the global optimal solution. Therefore, searching in dynamic multi-dimensional spaces presents a novel challenge for optimization algorithms.

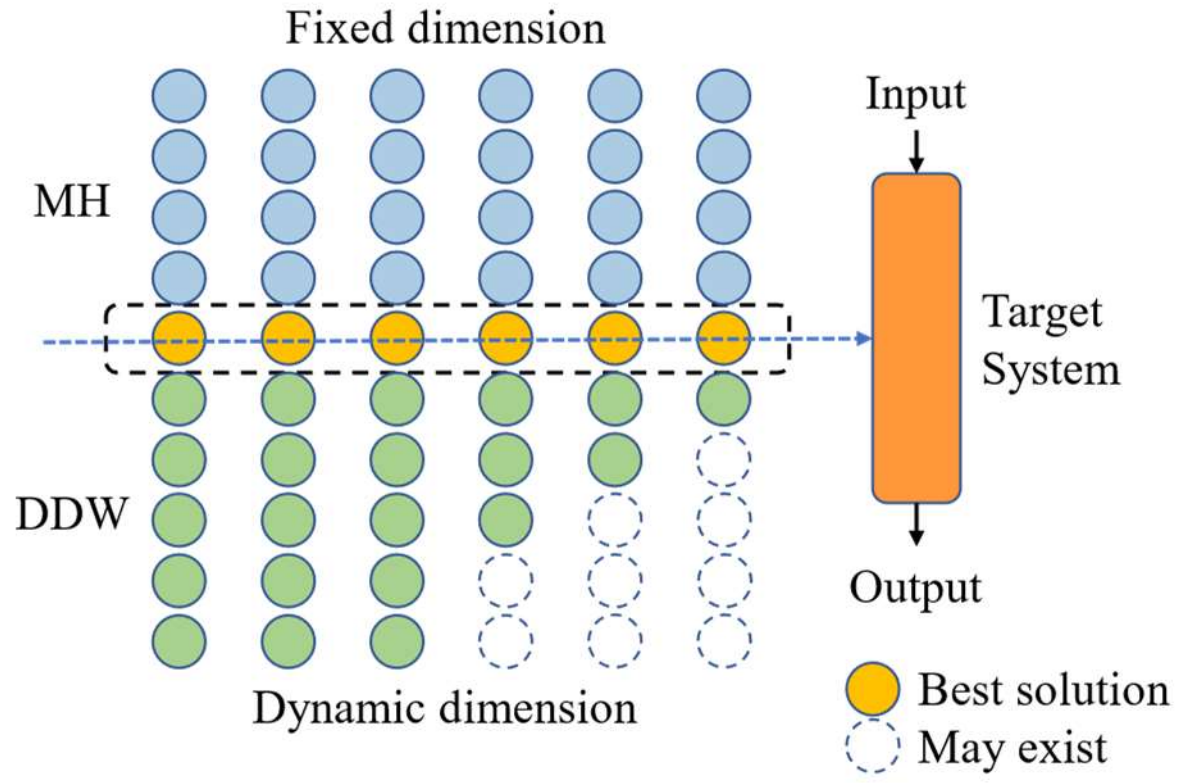

**Fig. 1.** The difference between MH and DDW

In order to search for optimal human motion templates in dynamic multidimensional space, we first established a dimension-crossing mapping mechanism and constructed an adaptability evaluation model for dynamic dimensionalization problems based on this mechanism. Subsequently, we proposed DDW, which possesses efficient cross-dimensional search capability. Finally, we compared and validated DDW against 31 metaheuristic algorithms. The results indicate that DDW exhibits a higher efficiency in dynamic multidimensional space search and can obtain superior human motion templates.

The remainder of this article is organized as follows. Section 2 introduces the experiment and dynamic multidimensional space. Section 3 presents the model of DDW. Section 4 introduces the experimental results. Section 5 discusses the results and methods. Finally, Section 6 concludes the paper.

## 2 Experiment and dynamic multidimensional space

### 2.1 Experiment

In this study, 6 healthy men (male, height: 1.81 ± 0.08 m, weight: 77.50 ± 9.5 kg, BMI: 23.81 ± 4.15) participated in the experimental data collection process. The subjects' postures (including back, thigh, and calf angle data) while walking on a Bertec treadmill (Bertec Corporation, Columbus, OH, USA) with an inclination angle of 0° at a speed of 4 km/h were captured using the Functional Assessment of Biomechanics (FAB, USA). The sampling frequency was 50 Hz. More than 80 continuous gait cycles were recorded for each participant. All subjects provided an informed consent form for the experiment, which was approved by the Ethics Committee at the Beijing Institute of Technology (approval code: BIT-EC-H-2024156).

### 2.2 Dynamic multidimensional space

The purpose of this study is to explore optimal temporal data chains that minimizes the distance to the original gait cycle data, thereby serving as reference templates for human motion. The specific process is as follows: After initializing $M$ random templates $X$, an iterative computation through the optimization algorithm is employed to identify an $x_k$, which minimizes the fitness to $N$ original gait cycles ($\theta$).

$$Minimize\ F(x_k, \theta) \tag{1}$$

$$\theta = \left(\theta_1, \dots, \theta_j, \dots, \theta_N\right) \tag{2}$$

$$\theta_j = \begin{cases} \theta_{back} = (\theta_{back}^1, \dots, \theta_{back}^i, \dots, \theta_{back}^D) \\ \theta_{l_thigh} = (\theta_{l_thigh}^1, \dots, \theta_{l_thigh}^i, \dots, \theta_{l_thigh}^D) \\ \theta_{r_thigh} = (\theta_{r_thig}^1, \dots, \theta_{r_thigh}^i, \dots, \theta_{r_thigh}^D) \\ \theta_{l_shank} = (\theta_{l_calf}^1, \dots, \theta_{l_calf}^i, \dots, \theta_{l_calf}^D) \\ \theta_{r_shank} = (\theta_{r_calf}^1, \dots, \theta_{r_calf}^i, \dots, \theta_{r_calf}^D) \end{cases} \tag{3}$$

$$X = \left(x_1, \dots, x_k, \dots, x_M\right) \tag{4}$$

$$x_k = \begin{cases} x_{back} = \left(x_{back}^1, \dots, x_{back}^i, \dots, x_{back}^{D1}\right) \\ x_{l_thigh} = \left(x_{l_thigh}^1, \dots, x_{l_thigh}^i, \dots, x_{l_thigh}^{D2}\right) \\ x_{r_thigh} = \left(x_{r_{thigh}}^1, \dots, x_{r_{thigh}}^i, \dots, x_{r_{thigh}}^{D3}\right) \\ x_{l_shank} = \left(x_{l_{calf}}^1, \dots, x_{l_{calf}}^i, \dots, x_{l_{calf}}^{D4}\right) \\ x_{r_shank} = \left(x_{r_{calf}}^1, \dots, x_{r_{calf}}^i, \dots, x_{r_{calf}}^{D5}\right) \end{cases} \tag{5}$$

$$D1, D2, D3, D4, D5 \in [D_{min}, D_{max}] \quad (6)$$

where, $\theta_j$ represents the data for the j-th gait cycle within $\theta$, encompassing five attributes corresponding to the back, left thigh, right thigh, left shank, and right shank. Each attribute of $\theta_j$ is an array with a dimensionality of $D$. $x_k$ represents the k-th template within $X$, which has the same attributes with $\theta_j$. However, the attributes of $x_k$ have varying dimensionalities $(D1, D2, D3, D4, D5)$.

It is noteworthy that, within $\theta$, the dimension quantity $D$ corresponding to different attributes of the same original gait cycle is equal. However, due to the incomplete symmetry of human motion, the dimension quantity $D$ of two different original gait cycles may not be equal, which results in $D$ being a non-constant value, with its statistical range falling between $[D_{min}, D_{max}]$. Therefore, $D1, D2, D3, D4, D5$ may be random integers within this range and are not necessarily equal to one another.

Figure 2 illustrates a dynamic multidimensional space composed of original gait data, which consists of multiple fixed-dimensional spaces (such as 59, 60, and 61). The number of fixed-dimensional spaces within the dynamic multidimensional space depends on $[D_{min}, D_{max}]$. In this configuration, the X-axis represents the gait cycle duration, the Z-axis represents the angular data, and the Y-axis denotes the dimensionality of the time data chain. The yellow and blue scatter points represent the sagittal plane angles of the right and left thighs, respectively. The optimal motion template may exist within any fixed-dimensional space, and optimal solutions for different attributes may also be distributed across multiple distinct fixed-dimensional spaces.

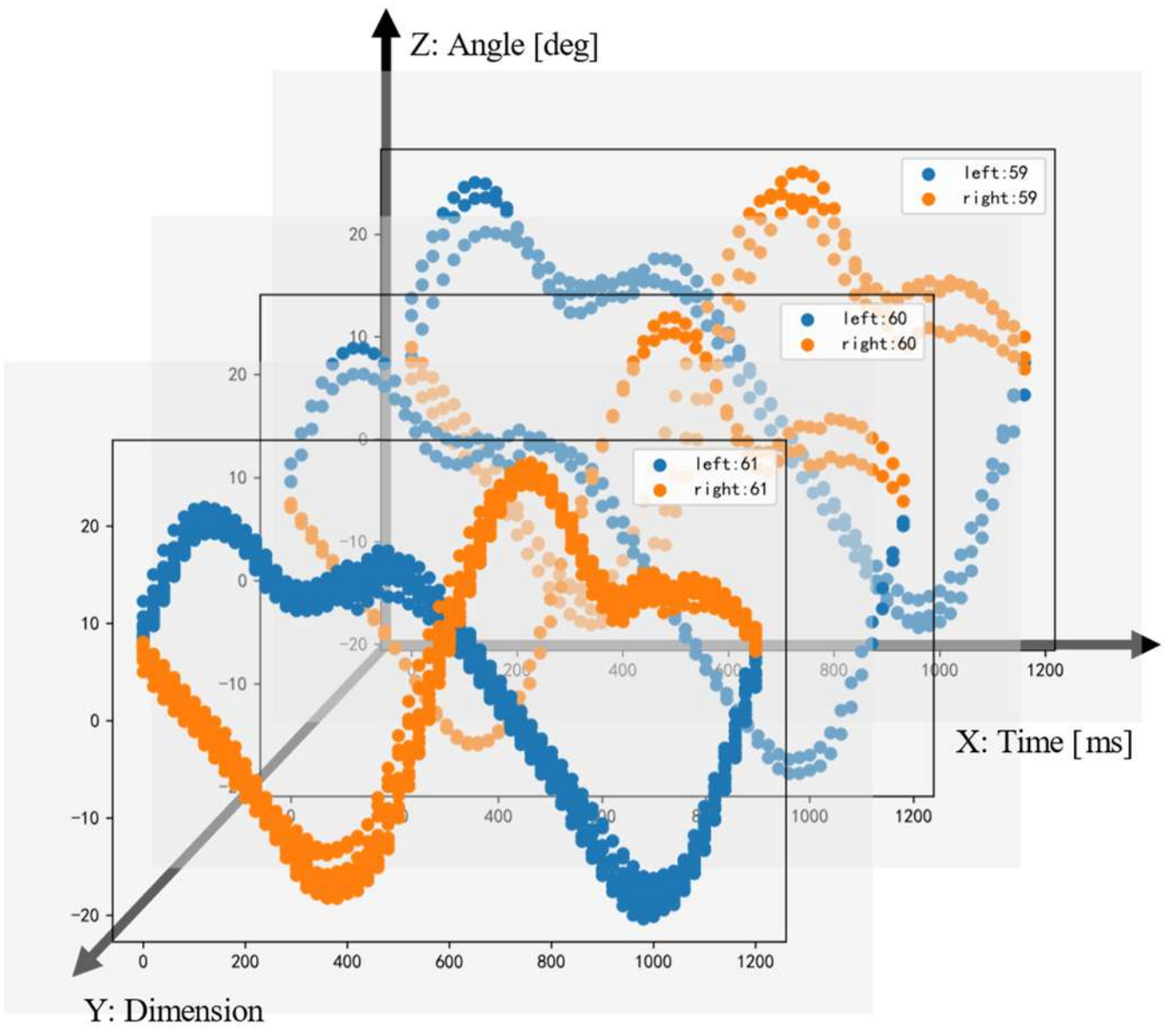

**Fig. 2.** Dynamic multidimensional space

Traditional optimization search algorithms, such as Particle Swarm Optimization (PSO), primarily focus on problems with a fixed number of dimensions. These methods are only applicable for target search within a single fixed multi-dimensional space. To cover the entire dynamic multi-dimensional space, it is necessary to conduct a global exhaustive search for each fixed multi-dimensional space, which is extremely inefficient and consumes substantial time and spatial resources. Furthermore, if traditional methods are limited to searching within a single fixed multi-dimensional space, they are likely to miss potential solutions present in other spaces, resulting in a tendency to become trapped in local optima and failing to achieve a global optimum. Therefore, the challenge of searching for optimal motion templates in dynamic multi-dimensional space lies in the need to perform distributed searches across different fixed multi-dimensional spaces. To address this, we introduce Dynamic Time Warping (DTW) technology, aiming to establish mapping relationships between data chains of various dimensions within the dynamic multi-dimensional space. Based on this, we propose a novel dynamic dimension warping optimization search algorithm to tackle the challenges presented by dynamic multi-dimensional space.

## 3 Dynamic Dimension Warping algorithm

The algorithmic flowchart of DDW is depicted in Fig. 3, which encompasses the following principal steps:

In Step 1, initialize the population ($X$) and compute the fitness of all individuals.

In Step 2, the necessary segments are selected to achieve optimal dimension gathering.

In Step 3, individuals within the population are stratified into three categories, labeled as A, B, and C, according to their fitness values. Novel individuals can then be synthesized using diverse strategies.

In Step 4, compute the fitness of all newly generated individuals and select a subset required to regenerate the population from both the newborn individuals and the existing population.

In Step 5, assess whether the termination criteria have been satisfied. If not, iterate steps 2 to 5; otherwise, output the optimal individual.

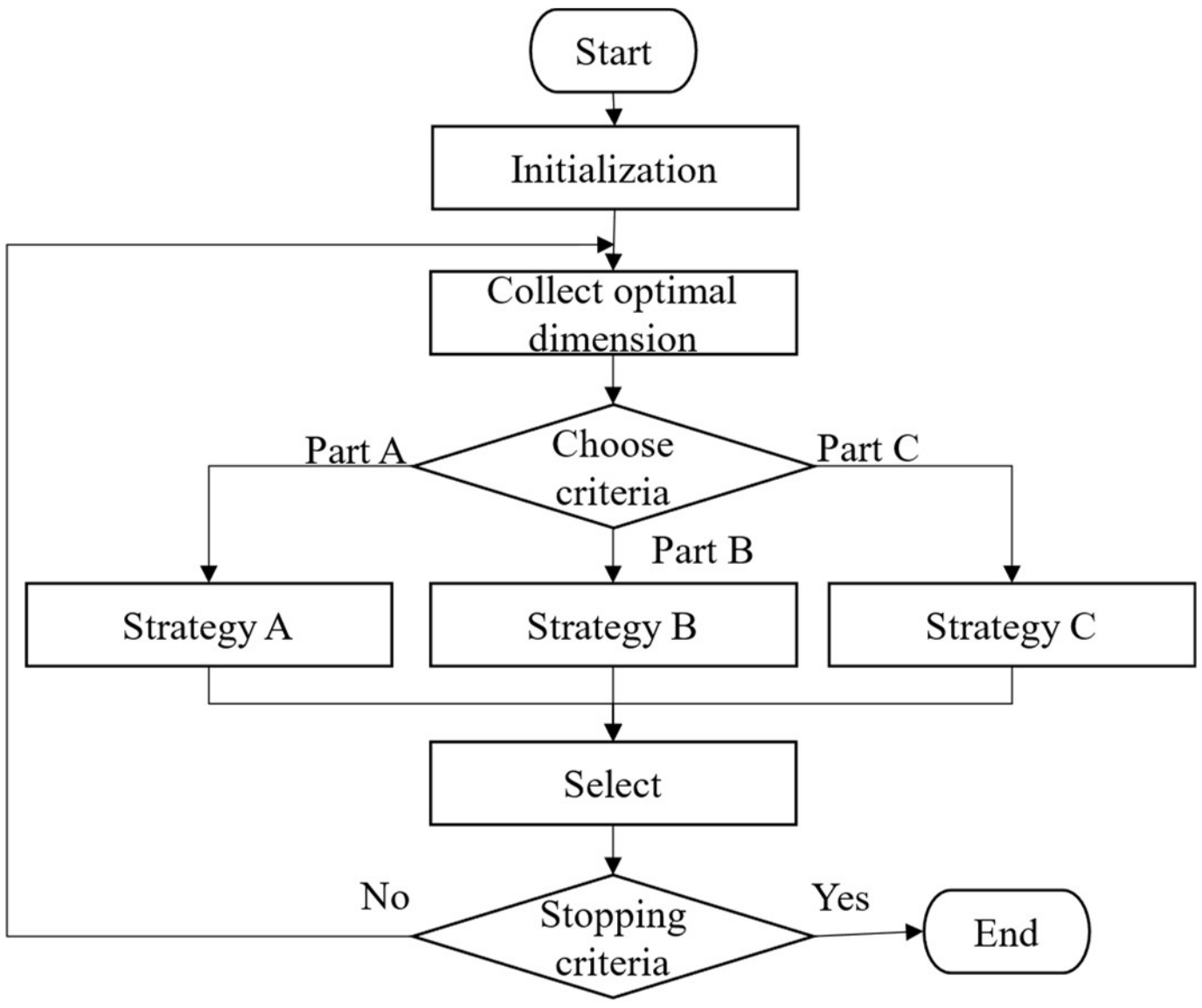

**Fig. 3.** The Flowchart of DDW

### 3.1 Initialization

Figure 4 illustrates the initialization process of "back" within $x_k$. The horizontal axis represents time and the vertical axis represents the angle. In the figure, the "Gaits" (blue scatter) represent the temporal data series of the gait cycle counts across various dimensions. The "Ave" (red line) denotes the average reference temporal data series, while the "Init" (green line) signifies the outcome of the ultimate initialization process.

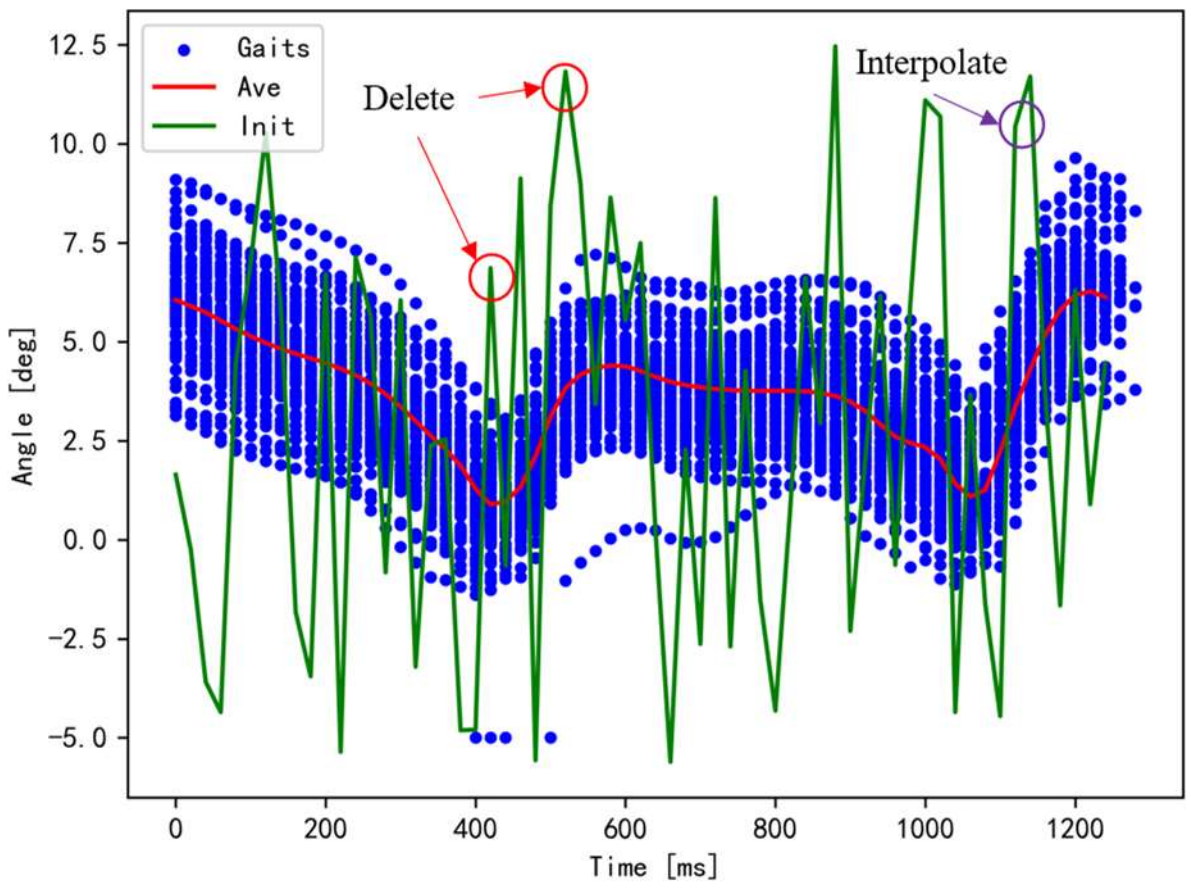

**Fig. 4.** Initialize the 'back' of $\boldsymbol{x_k}$

$$D_{count} = Mode(D_1, \dots, D_j, \dots, D_N), D_j = D\ of\ \theta_j \tag{7}$$

$$\theta_{ave-back} = (\theta_{back}^1, \dots, \theta_{back}^i, \dots, \theta_{back}^{D_{count}}) \tag{8}$$

$$\theta_{ave-back}^i = \sum_{j=1}^{count} \theta_{j-back}^i / count\ , \theta_j \in \theta, D\ of\ \theta_j = D_{count} \tag{9}$$

$$D1 = randint(D_{min}, D_{max}) \tag{10}$$

$$x_{k-back}^i = \theta_{ave-back}^i + rand(\theta_{back}^{min}, \theta_{back}^{max}) \tag{11}$$

$D_{count}$ represents the mode of the dimensionality count for each gait cycle in $\theta$ (as shown in "Gaits" in Fig. 3), and $count$ represents the statistical count of gait cycles with dimensionality equal to $D_{count}$; $\theta_{ave-back}$ represents the mean value of all gait cycles with dimensionality equal to $D_{count}$ in each dimension (Equations 12 and 13); $randint$ means to get a random integer inside $[D_{min}, D_{max}]$, while $D1$ represents the dimensionality count of $x_{k-back}$; $\theta_{back}^{min}$ and $\theta_{back}^{max}$ represent the minimum and maximum values of attribute "back" in each dimension, respectively. As shown in "Init" in Fig. 3, to ensure the final dimensionality count is $D1$, we must randomly delete some data or perform linear interpolation on random portions.

The other attributes of $x_k$ are then initialized using the same method. This process will be repeated M times to complete the initialization of the entire population. Each individual in the population is a particle in the multidimensional space, and each dimension of each attribute describes the position of the particle in the multidimensional space. Each individual is also a potential solution for practical problems.

### 3.2 Fitness function

To compute the fitness value of each individual in the population, we need to consider the multidimensional time series attributes of each individual (e.g., A and B, see Formula 12). Owing to the potentially different dimensionalities of these time series, we employed methods based on DTW and Euclidean distance (Formulas 13-15) to establish mappings between different dimensions.

$$A = (a_1, a_2 \dots, a_i, \dots, a_{l_A}), B = (b_1, b_2 \dots, b_i, \dots, b_{l_B}) \tag{12}$$

$$V, v, dir = Map(A, B) = \begin{cases} DTW(A, B), & l_A \neq l_B \\ Euclidean(A, B), & l_A = l_B \end{cases} \quad (13)$$

$$dir = (dir_1, dir_2 \ldots, dir_i, \ldots, dir_{l_A}) \quad (14)$$

$$v = (v_1, v_2 \ldots, v_i, \ldots, v_{l_A}), v_i = min\left(\left\{(a_i - b_j)^2 \middle| j \in dir_i\right\}\right) \quad (15)$$

A and B can represent time series which can be any attribute of any element in $\theta$ or $X$. $V$ represents the distance between A and B; $v$ represents the distance of the corresponding elements mapped from B to A; $dir$ represents the set of indices of the corresponding elements mapped from B to A, which also indicates the mapping direction from B to A.

As shown in Fig. 5, when $l_A = l_B$, both $v$ and $dir$ are the results of each element in A and B being one-to-one correspondence. However, when $l_A \neq l_B$ occurs, we need to search for the optimal path through the path matrix in the DTW calculation process.

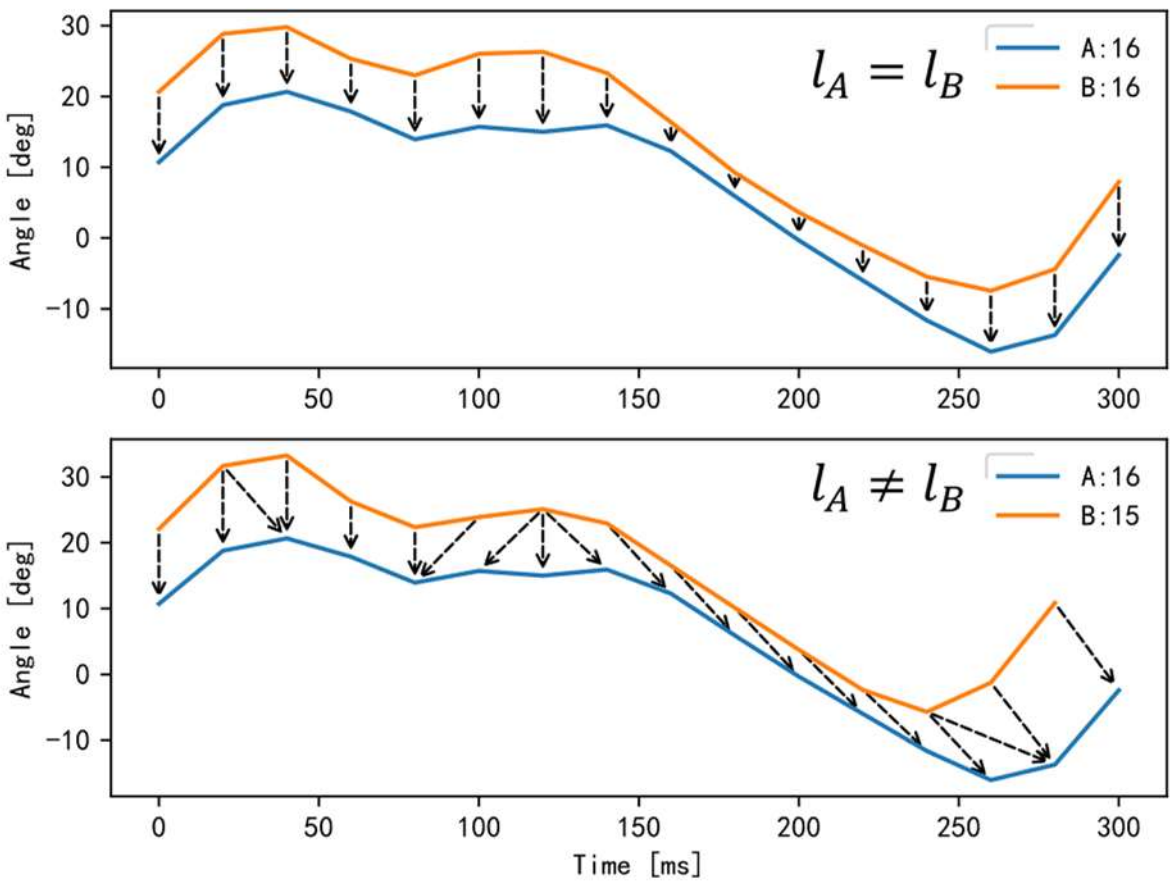

**Fig. 5.** Mapping from B to A

Fig. 6 shows the computational details of the DTW. "Dis" and "Route" denote the distance matrix and path matrix generated during the DTW computation, respectively. $Route[l_A - 1, l_B - 1]$ denotes the endpoint of the path matrix, i.e., the direct distance $V$ between both A and B (Equation 13). "Best Route" denotes the set of consecutive elements in the distance matrix that makes $V$ the minimization. In the optimal path, $a_i$ may have a mapping relationship with multiple elements in B, which means that $dir_i$ is a set containing multiple element indices in B. Simultaneously, $v_i$ reflects the minimum distance between $a_i$ and these multiple elements in B.

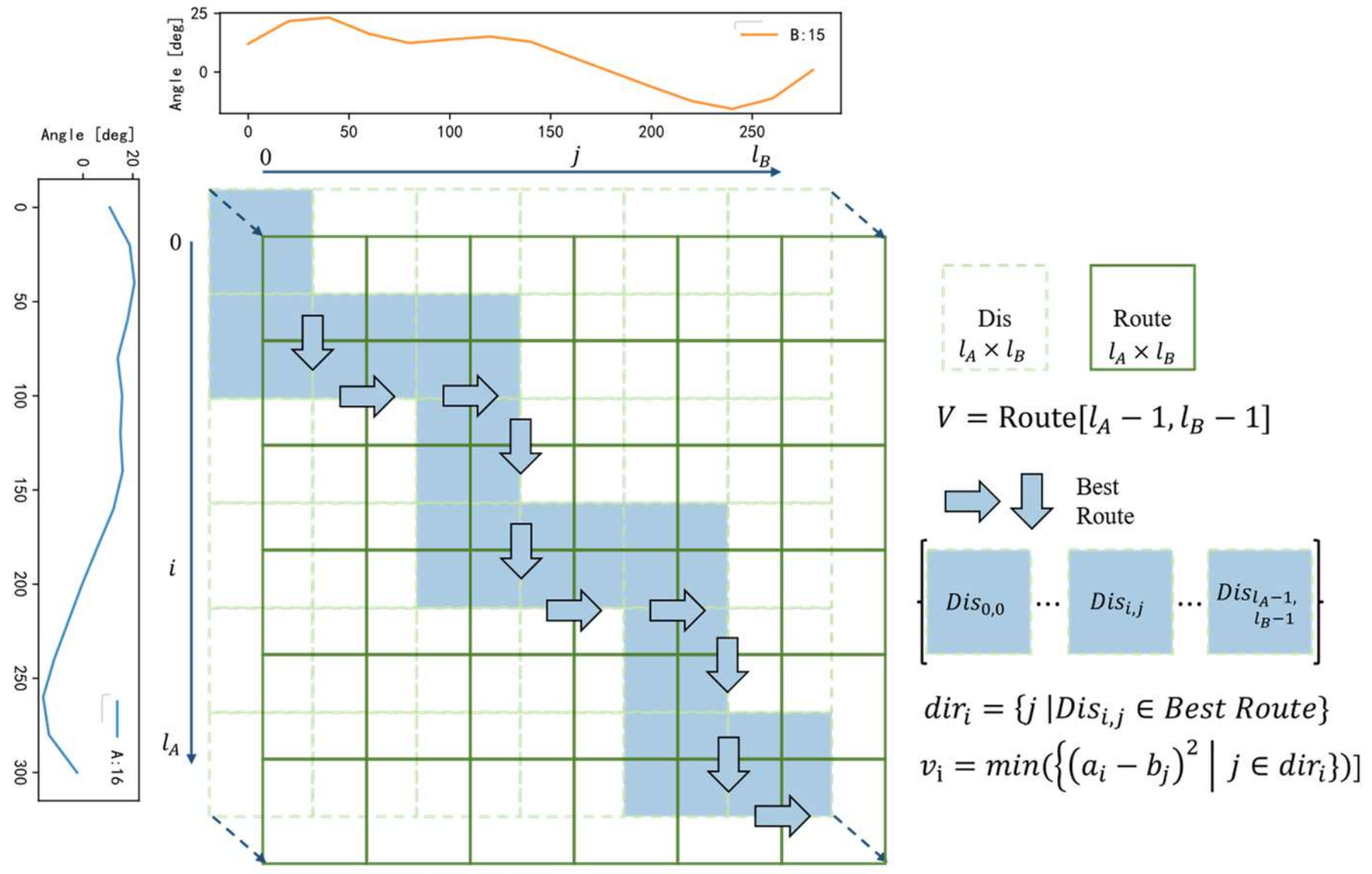

**Fig. 6.** Details of DTW

Based on the calculation method above, the process of obtaining the fitness $fit_k$ of $x_k$ is as follows:

$$fit = \{fit_1, \dots, fit_k, \dots, fit_M\} \quad (16)$$

$$fit_k = F(x_k, \theta) = \sum_{j=1}^{N} f(x_k, \theta_j)/N \quad (17)$$

$$f(x_k, \theta_j) = Ave(\{V_{back}, V_{l_thigh}, V_{r_thigh}, V_{l_calf}, V_{r_calf}\}) \quad (18)$$

$fit$ represents the set of fitness values of all individuals in the entire population; $f(x_k, \theta_j)$ represents the distance between $x_k$ and $\theta_j$; $Ave()$ represents the average value of all elements in the set; $\{V_{back}, V_{l_thigh}, V_{r_thigh}, V_{l_calf}, V_{r_calf}\}$ represents the distance between $x_k$ and $\theta_j$ on various attributes, which is calculated by Formula 17.

As depicted in Fig. 7, following each update of $X$ and $fit$, the population undergoes partitioning into "Part A" (5%), "Part B" (45%), and "Part C" (50%) based on the ascending order of fitness values. These three sections will generate new individuals with diverse strategies. $x_{best}$ represents the least fit individual in the current population.

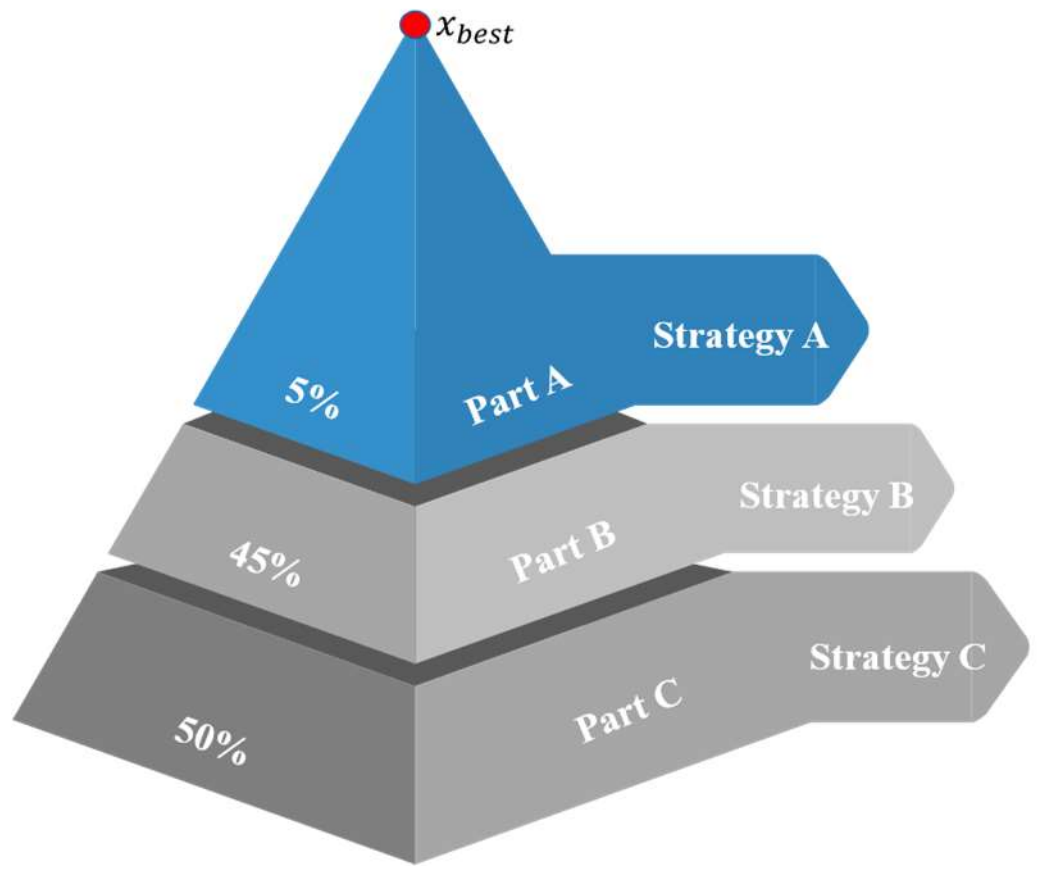

**Fig. 7.** Population division

### 3.3 Optimal Dimension Collection

Optimal Dimension Collection (ODC) involves establishing inter-dimensional mapping relationships with all other individuals in "Part A" based on $x_{best}$ (see Formula 17), aiming to determine the optimal dimension value for each dimension within the best group of the current population. The specific process is as follows：

$$D_{best-back} = \{D_{best-back}^{1}, \dots, D_{best-back}^{i}, \dots, D_{best-bac}^{len(x_{best-bac})}\} \tag{19}$$

$$V, v, dir = Map(x_{best-bac}, x_{j-back}) \tag{20}$$

$$D_{best-back}^{i} = \begin{cases} x_{best}^{dir-i}, if\ v_{x_{best}}^{dir-i} \le v_{x_j}^{dir-i} \\ x_j^{dir-i},, if\ v_{x_{best}}^{dir-i} > v_{x_j}^{dir-i} \end{cases} \tag{21}$$

$D_{best-back}$ represents the "back" attribute of the Optimal Dimension Solution (ODS) $D_{best}$. $x_j$ is a randomly selected solution ranked j-th in $X$ based on fitness, but it cannot be itself. $v$ and $dir$ provide mapping information between $x_{best-back}$ and $x_{j-b}$ after dimension folding. $dir - i$ represents the dimension in $x_{j-b}$ corresponding to the i-th dimension in $x_{best-back}$. $v_{x_{best}}$ and $v_{x_j}$ respectively denote the minimum dimensional distances obtained by $x_{best-back}$ and $x_{j-back}$ with respect to each dimension corresponding to all elements in $\theta$ during the fitness calculation process.

Following the aforementioned method, the remaining attributes of $D_{best}$ are resolved. Building upon this, we substitute $D_{best}$ for $x_{best}$ and proceed to explore, alongside other individuals in "Part A", the optimal value for each dimension, ultimately deriving the definitive $D_{best}$.

### 3.4 Strategy A

In order to avoid getting trapped in local optimization and search for better positions globally, individual $x_a$ in "Part A" randomly moves around individual $x_{best}$. The movement process is as follows:

$$x_{a-new-back}^{j} = x_{best}^{j} + (\theta_{back}^{max} - \theta_{back}^{min}) * step \tag{22}$$

$$step = Lévy(\lambda) * (1 - (gen/Max_gen)^2) \tag{23}$$

$$D_{new} = randint(D_{min}, D_{max}) \tag{24}$$

$x_{a-new-back}^{j}$ represents the value of the j-th dimension of the "back" attribute of the new individual $x_{a-new}$ for $x_a$ (Equation 22). Lévy(λ) generates a random number based on the heavy-tailed power-law step size distribution, which has infinite variance and mean. λ is a parameter within the interval (1,3]. $iter$ represents the current iteration number, while $Max_iter$ represents the maximum iteration number. $D_{new}$ is the required number of dimensions.

To ensure that the final number of dimensions is $D_{new}$, multiple deletions of the worst dimension data or linear interpolation near the worst dimension need to be performed on $x_{a-new-back}$. The worst dimension refers to the dimension with the maximum dimensional distance (Equation 13).

Based on the process above, the solution for the other attributes of the new individual $x_{a-new}$ will be completed.

### 3.5 Strategy B

In "Part B," the current individual $x_b$ centers around its own position, utilizing information from other individuals, and explores three paths using the Archimedean spiral method, thereby generating three new individuals. The search process is as follows:

$$x_{b-new-back}^{j} = x_{b-back}^{j} + \begin{cases} x_{b-bac}^{j} & + route_1 \\ x_{b-b}^{j} & + route_2 \\ x_{b-bac}^{j} & + route_3 \end{cases} \tag{25}$$

$$route_1 = X(j) * \left(x_{better-bac}^{dir_j} - x_{b-bac}^{j}\right) \tag{26}$$

$$route_2 = Y(j) * (D_{b-bac}^{dir_j} - x_{b-bac}^{j}) \tag{27}$$

$$route_3 = route_1 + route_2 \tag{28}$$

$$X(j) = \frac{xr(j)}{\max(|xr|)}, Y(j) = \frac{yr(j)}{\max(|yr|)} \tag{29}$$

$$xr(j) = r(j) * \sin(\theta(j)), yr(j) = r(j) * \cos(\theta(j)) \tag{30}$$

$$r(j) = \theta(j) + 1.5 * rand(0,1), \theta(j) = 10\pi * rand(0,1) \tag{31}$$

$x_{b-new-back}^{j}$ represents the value of the j-th dimension of the "back" attribute of the new individual $x_{b-new}$ for $x_b$. $x_{better-bac}^{dir_j}$ represents the value of the $dir_j$-th dimension of the attribute "back" of $x_{better}$ corresponding to $x_{b-back}^{j}$. $x_{better}$ is an individual with a fitness value less than $x_b$ randomly selected from "Part A" and "Part B". $D_{b-bac}^{dir_j}$ represents the value of the $dir_j$-th dimension of the attribute "back" of $D_{best}$ corresponding to $x_{b-back}^{j}$. $dir$ can be calculated using equation 13.

Based on the above process, the solution for other attributes of the new individual $x_{b-new}$ will be completed.

### 3.6 Strategy C

As “Part C” constitutes the largest proportion of the population, it significantly influences the overall search efficiency. Consequently, individuals in this group (such as $x_c$) adopted a multi-path competitive search strategy centered around both the group’s best individual and the best individual in each dimension during the updating process. In contrast to “Part B,” which represents a relatively underperforming group, “Part C” has remained on the periphery of the population for an extended period, necessitating a rapid convergence toward optimal individuals. Thus, during the search process, in addition to focusing on the optimal individuals, $x_c$ also introduced a new search direction known as ODS. The search process is as follows:

$$x_{c-new-bac}^{j} = \begin{cases} x_{best-bac}^{j} & + route_1 \\ D_{best-back}^{j} & + route_2 \\ x_{best-bac}^{j} & + route_3 \end{cases} \tag{32}$$

$$route_1 = X1(j) * (x_{c-bac}^{j} - x_{best-bac}^{dir1_j}) \tag{33}$$

$$route_2 = Y1(j) * (x^{j}_{c-bac} \; - D^{dir2_j}_{best-bac} \; ) \tag{34}$$

$$route_3 = route_1 + route_2 \tag{35}$$

$$X1(j) = \frac{xr(j)}{max\ (|xr|)}, Y1(j) = \frac{yr(j)}{max\ (|yr|)} \tag{36}$$

$$xr(j) = r(j) * sinh(\theta(j)), yr(j) = r(j) * cosh(\theta(j)) \tag{37}$$

$$r(j) = \theta(j) + 1.5 * rand(0,1), \theta(j) = 10\pi * rand(0,1) \tag{38}$$

$x^{j}_{c-new-back}$ represents the value of the j-th dimension of the "back" attribute of the new individual $x_{c-new}$ for $x_c$. $D^{j}_{best-back}$ and $x^{j}_{best-back}$ represent the values of the j-th dimension of the "back" attribute for $D_{best}$ (ODS, Equation 19) and the optimal individual $x_{best}$, respectively.

Based on the aforementioned process, the solution for the other attributes will be completed.

### *3.7 Selection*

After completing the update calculations for all individuals, it is necessary to select the desired individuals based on their fitness values to recombine a new population for the next round of computations. As shown in Fig. 8, the new population consists of two parts. All the new individuals generated by "Part B" and "Part C", along with the original population $X$ and the dimensionally optimal individual $D_{best}$, will form a collective set. The first part will select $95\% * M$ individuals with relatively lower fitness values from this collective set. The second part will directly consist of new individuals generated from "Part A", accounting for 5% of the population.

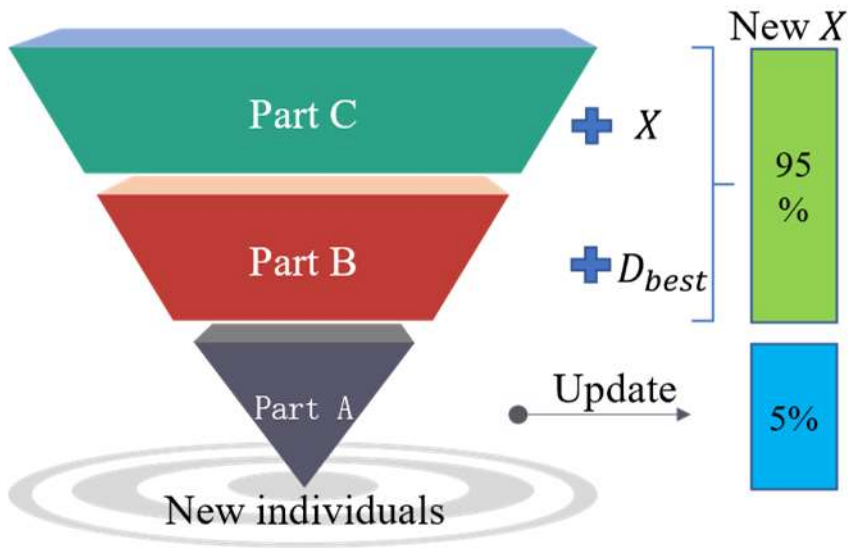

**Fig. 8.** The process of population renewal

## 4 Experimental results

The algorithms were programmed in Python (3.10.5) and executed on computation environment of Intel Core i7-7700HQ CPU 2.80 GHz, 2.5 GHz, 16 GB RAM and 64-bit operating system.

### 4.1 Test for ODC

The ODC constitutes a pivotal stage within DDW, with the ODS generated therein exerting a significant influence on the search and updating processes of the entire population. To evaluate its ramifications, we conducted 1,000 tests of ODC for two randomly selected individuals at the culmination of each population iteration. In this test, the population size of the algorithm and the maximum number of iterations were set to 50 and 500.

Fig. 9 illustrates the test results of ODC, depicting the fitness variations of the ODS throughout the entire iteration process. "Best" and "Better" denote the likelihood of surpassing all and some of the original individuals, respectively. "Worst" indicates the probability of performing worse than all the original individuals. "Ave", "Std", and "Min" denote the mean, standard deviation, and minimum of the fitness values, respectively.

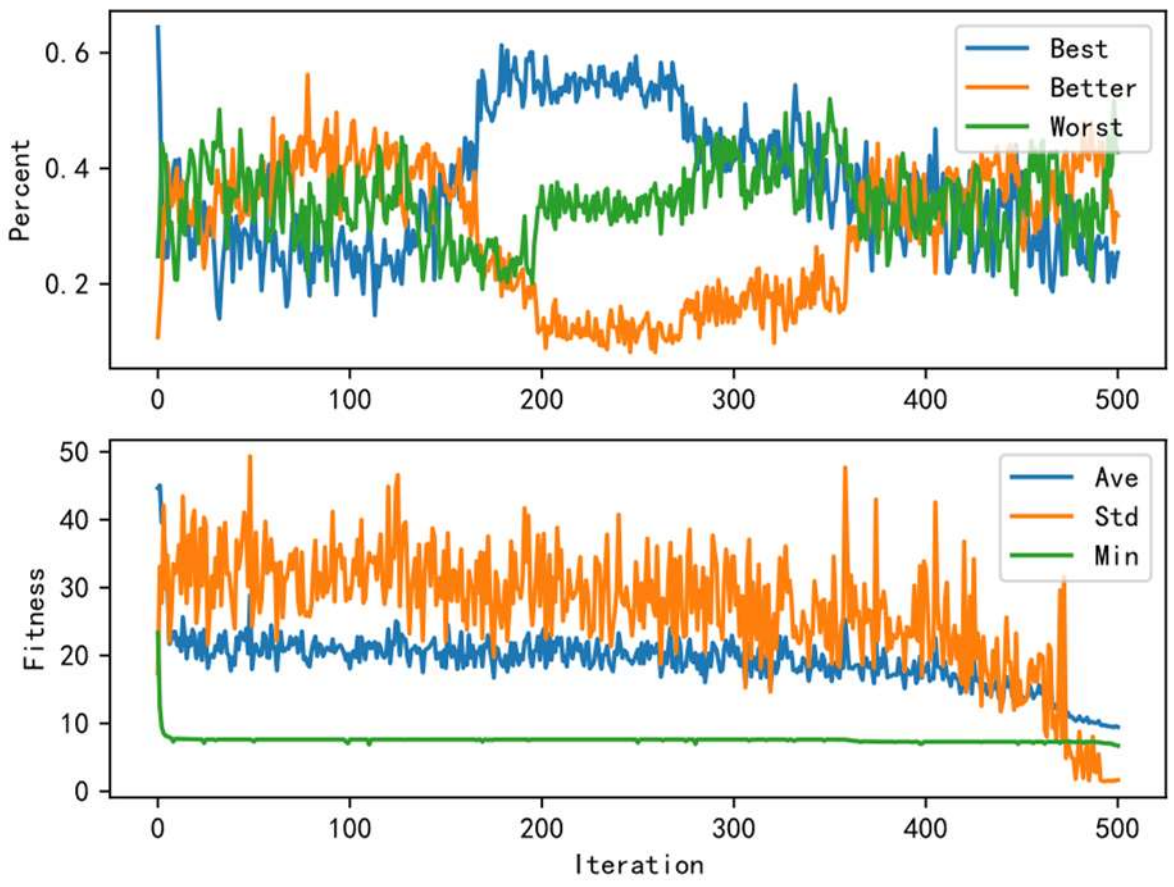

**Fig. 9.** Results of ODC test

Throughout 500 iterative searches, the global average probabilities of "Best", "Better", and "Worst" collected by the optimal dimension are 37.23%, 28.68%, and 34.09%, respectively. This indicates that this unique search method yielded superior individuals more than 65% of the time. Within the initial six iterations, the probability of "Best" collected from the optimal dimension exceeded 40%. Simultaneously, the minimum value of individual fitness was less than 10 by the third iteration, aligning with the ideal value set in this study. By the fifth iteration, within the initial ten iterations, the average individual fitness approached 20, signifying the transition of the group search into the refinement stage.

The findings from this test highlight the crucial role of ODC throughout the entire search process. It significantly narrows down the search space and is anticipated to converge towards the true optimal solution. Moreover, the average fitness values of individuals hovered around 20 during the mid-term search phase and swiftly approached the minimum feasible value in subsequent iterations. Hence, the ODC furnishes essential support for the population-wide search across various iteration stages.

### *4.2 Motion template construction*

To assess the motion template construction capability of DDW, we conducted comparative experiments employing 31 meta-heuristic algorithms. For this assessment, the population size and maximum number of iterations for all algorithms were uniformly set to 50 and 500, respectively. The critical parameters of all algorithms are delineated in Table 1.

**Table 1.** Parameter values of DDW and other algorithms

| Algorithm | Parameter | Value |
|---|---|---|
| Fireworks Algorithm (FWA) [22] | $a, b$ | 0.2, 0.8 |
| Genetic Algorithm (GA) [12] | - | - |
| Artificial Bee Colony Algorithm (ABC) [23] | $Fs, limit$ | 10, 60 |
| Cuckoo search (CS) [24] | $pa, \alpha$ | 0.3, 1 |
| Firefly Algorithm (FA) [25] | $\alpha, \beta_{max}, \beta_{min}, \gamma$ | 0.97, 1.0, 0.2, 1 |
| Group Search Optimizer (GSO) [26] | $l_{max}$ | 3 |
| Water wave optimization (WWO) [18] | $h_{max}, \lambda, \alpha, \beta$ | 5, 0.5, 1.0026, 0.0001 |
| Bat Algorithm (BA) [27] | $\alpha, \gamma, f_{max}, f_{min}$ | 0.85, 0.9, 1, 0 |
| Shuffled Frog Leaping Algorithm (SFLA) [28] | $m$ | 10 |
| Gravitational Search Algorithm (GSA) [16] | $G_0$ | 100 |
| Grey Wolf Optimizer (GWO) [8] | $a$ | 2 |
| Brain Storm Optimization (BSO) [29] | $k, p5a, p6b, p6b3, p6c$ | 20, 0.2, 0.8, 0.4, 0.5 |
| Sparrow Search Algorithm (SSA) [30] | $ST, PR, SD$ | 0.8, 80%, 20% |
| Ant Lion Optimization (ALO) [31] | - | - |
| Butterfly Algorithm (BFA) [32] | $p, c, a$ | 0.8, 0.001, 0.1 |
| Monarch Butterfly Optimization (MBO) [33] | $p, BAR, S_{max}, peri$ | 5/12, 5/12, 100, 1.2 |
| Moth-Flame Optimization (MFO) [9] | $t$ | -1 |
| Harmony Search (HS) [19] | $HMCR, PAR$ | 0.9, 0.1 |
| Mayfly Optimization Algorithm (MA) [34] | $a_1, a_2, \beta, fl, d$ | 1, 1.5, 2, 0.1, 0.1 |
| Grasshopper Optimisation Algorithm (GOA) [35] | $f, l, C_{max}, C_{min}$ | 0.5, 1.5, 1e-5, 1 |
| Bald Eagle Search (BES) [36] | $c_1, c_2, \alpha, a, R$ | 2, 2, 2, 10, 1.5 |
| Marine Predators Algorithm (MPA) [37] | $FADs$ | 0.2 |
| Archimedes Optimization Algorithm (AOA) [17] | $C_1, C_2, C_3$ | 2, 6, 2 |
| Salp Swarm Algorithm (SSA-2) [38] | - | - |
| Slime Mould Algorithm (SMA) [39] | $Z$ | 0.03 |
| Pigeon-inspired Optimization (PIO) [40] | $NcRate, R$ | 0.75, 0.2 |
| Aquila Optimizer (AO) [10] | - | - |
| Harris Hawks Optimization (HHO) [41] | - | - |
| Dingo Optimization Algorithm (DOA) [11] | $P, Q$ | 0.5, 0.7 |
| Whale Optimization Algorithm (WOA) [42] | $a, b$ | 2, 1 |
| Particle Swarm Optimization (PSO) [7] | $\omega, C_1, C_2$ | 0.8, 1.49445, 1.49445 |
| DDW | - | - |

Algorithms other than DDW lack the capability to traverse dimensions, necessitating uniformity in the number of dimensions across attributes for all individuals within their respective populations. To facilitate comparison and construction of optimal templates with varying dimensionalities, we initialized three distinct populations, setting the initial dimensions to correspond to the top three values in the dimension quantity statistics. The initialization process for these comparison algorithms closely mirrors that of DDW, albeit with fixed dimensionality for all attributes of all individuals, unlike DDW where the dimensionality of any attribute for any individual is randomly selected within the statistical range. Subsequently, each algorithm underwent a minimum of three repeated tests to mitigate the impact of random variations in algorithmic calculations.

Table 2 presents the average (Ave), minimum (Min), standard deviation (Std), and overall ranking (Rank) of fitness for the optimal motion templates discovered by 32 algorithms. The overall ranking is primarily based on the evaluation metrics "Ave" and "Min." Among these metrics, the five algorithms that performed the best are DDW, BES, MA, MBO, and FA. DDW demonstrates superior performance in terms of both mean and minimum values. The average fitness value of the motion templates generated by DDW is 9.16, marking a remarkable 41% reduction compared with the overall average across all 32 algorithms. This indicates that DDW has the highest possibility of searching for the motion template closest to the actual optimal solution. Notably, the minimum fitness value attained by DDW stands at 8.54, representing a substantial 37% decrease relative to the average fitness level observed across all 32 algorithms. This assessment underscores each algorithm's efficacy in achieving optimal solutions within finite temporal and spatial constraints. Moreover, multiple repeated tests eliminate the influence of randomness on the search process of all algorithms to some extent. Given the practical utility of optimal computational outcomes, this analysis focuses solely on the minimum fitness values, omitting consideration of the maximum values. Furthermore, the standard deviation serves as a measure of the variability inherent in the computational outcomes of each algorithm. DDW's standard deviation in fitness calculations ranks 7th among the 32 algorithms assessed, indicating a notable 65% reduction compared to the average standard deviation observed across all algorithms. In summary, compared with other algorithms, DDW can stably search for the optimal motion template with the highest possibility, and the searched motion template is closest to the actual optimal solution.

**Table 2.** Test results for motion template construction

| Algorithm | Ave | Min | Std | Rank | Algorithm | Ave | Min | Std | Rank |
|---|---|---|---|---|---|---|---|---|---|
| FWA | 21.58 | 19.03 | 1.77 | 28 | MFO | 21.45 | 18.87 | 1.66 | 27 |
| GA | 16.99 | 14.60 | 1.57 | 21 | HS | 9.64 | 9.07 | 0.35 | 6 |
| ABC | 10.69 | 9.90 | 0.51 | 11 | MA | 9.43 | 8.88 | 0.40 | 2 |
| CS | 22.74 | 20.06 | 1.84 | 31 | GOA | 12.20 | 10.60 | 1.03 | 13 |
| FA | 9.59 | 9.06 | 0.39 | 5 | BES | 9.45 | 8.88 | 0.40 | 3 |
| GSO | 21.55 | 18.00 | 2.04 | 23 | MPA | 20.90 | 18.44 | 1.78 | 24 |
| WWO | 13.27 | 11.41 | 1.01 | 14 | AOA | 10.24 | 9.56 | 0.44 | 8 |
| BA | 21.99 | 18.75 | 2.02 | 26 | SSA-2 | 15.44 | 13.69 | 1.19 | 18 |
| SFLA | 10.58 | 9.81 | 0.47 | 10 | SMA | 14.39 | 12.55 | 1.09 | 16 |
| GSA | 22.77 | 19.58 | 1.91 | 29 | PIO | 18.78 | 16.54 | 1.52 | 22 |
| GWO | 16.05 | 14.02 | 1.45 | 19 | AO | 10.26 | 9.58 | 0.43 | 9 |
| BSO | 16.30 | 14.25 | 1.36 | 20 | HHO | 9.73 | 9.10 | 0.36 | 7 |
| SSA | 13.59 | 11.45 | 1.39 | 15 | DOA | 21.26 | 18.70 | 1.72 | 25 |
| ALO | 22.74 | 20.08 | 1.89 | 32 | WOA | 11.04 | 10.13 | 0.52 | 12 |
| BFA | 22.71 | 19.63 | 1.97 | 30 | PSO | 17.08 | 13.08 | 2.64 | 17 |
| MBO | 9.58 | 8.98 | 0.38 | 4 | DDW | 9.16 | 8.54 | 0.42 | 1 |

The fluctuation in the fitness level of the top individual over the iterative process serves as an indicator of the search efficiency within the population. Drawing from the insights gleaned from Table 2, Fig. 10 depicts variations in average fitness, minimum fitness of the top individual, and standard deviation of population fitness across iterations for the top 5 algorithms (DDW, BES, MA, MBO, FA).

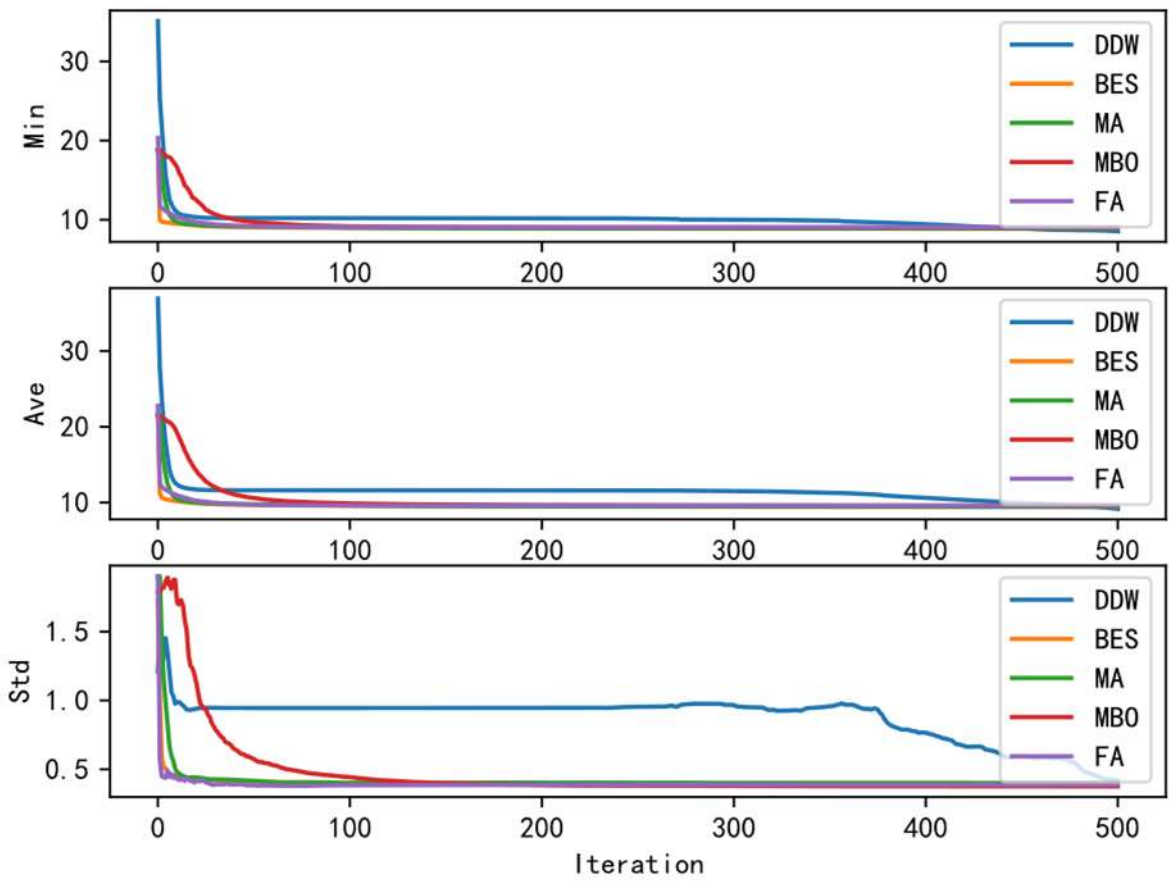

**Fig. 10.** The iterative process of the top 5 algorithms

The five algorithms demonstrate remarkable search efficiency in the initial exploration phases. By the 10th iteration, the average fitness of the best individual is significantly lower than that of other algorithms, demonstrating its ability to quickly locate potential optimal solutions. Throughout the search process, DDW maintained a relatively high level of population fitness standard deviation (approximately 0.9), indicating its ongoing effective exploration in multiple directions rather than a singular focus. In contrast, other algorithms exhibited lower standard deviations (approximately 0.4), suggesting a more concentrated exploration along a specific direction. Consequently, DDW did not manifest a discernible advantage during the initial stages of the search, merely demonstrating search efficiency comparable to that of other algorithms.

Although DDW maintained a high level of exploratory freedom in the early stages of the search, it demonstrated remarkable performance towards the end of the search process by rapidly converging its population to the optimal solution. Furthermore, the final search results achieved by DDW surpassed those of all other algorithms. This exceptional performance can be attributed to DDW's ability to effectively balance exploration and exploitation strategies throughout the search process. In summary, the DDW algorithm excels in maintaining population diversity and achieving rapid convergence, thereby endowing it with a significant advantage in addressing dynamic problems.

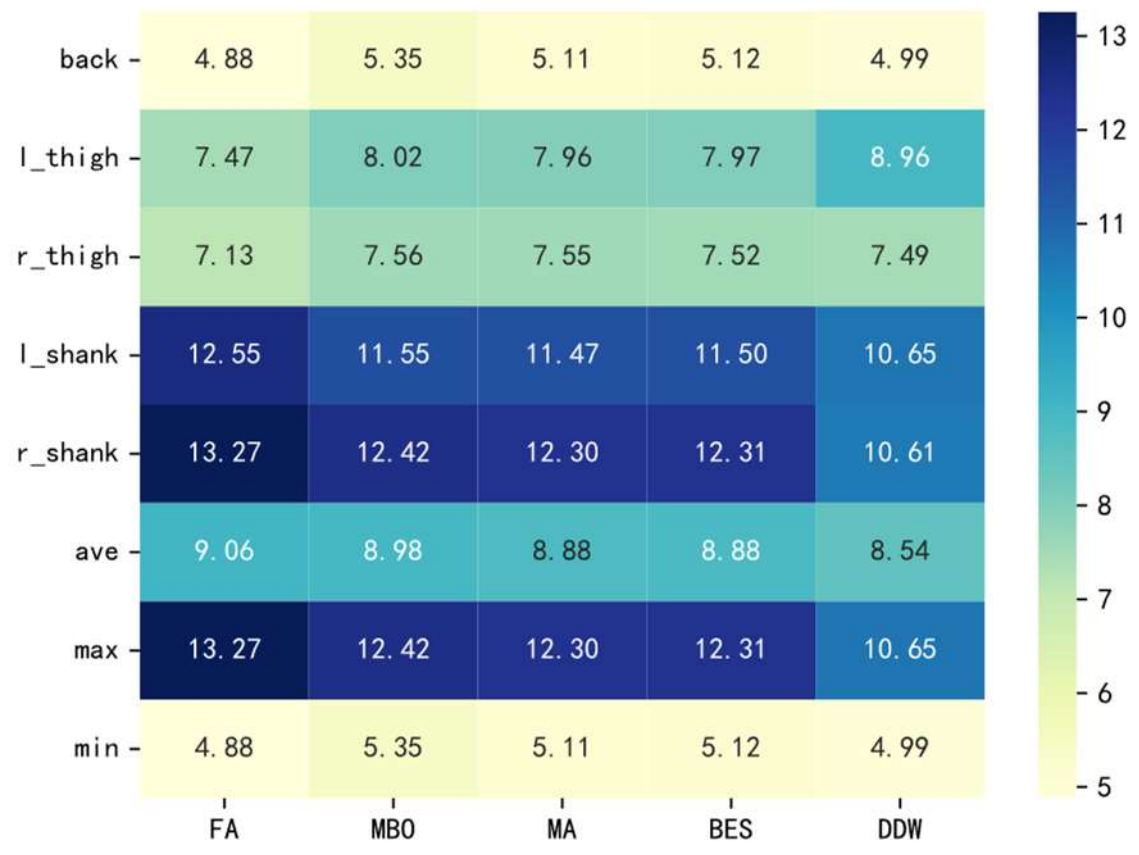

**Fig. 11.** Search results for Motion templates

Figure 11 further illustrates the performance of five leading algorithms (FA, MBO, MA, BES, DDW) in the optimal motion template search task, and provides a detailed summary of their fitness values in the back, left thigh, right thigh, left shank, and right shank (back, l_thigh, r_thigh, l_shank, r_shank). Although the performance of the DDW algorithm in specific attributes such as "back," "l_thigh," and "r_thigh" is comparable to that of the other algorithms, considering the strong coupling among attributes within the motion template, we emphasize the evaluation of overall performance. Under this evaluation criterion, defined as average performance (ave), the DDW algorithm stands out with the lowest average value of 8.54, demonstrating its ability to consistently find solutions close to the optimal during the overall search process. Furthermore, in the assessment of best performance (min) and worst performance (max), the DDW algorithm also exhibits exceptional robustness, with its minimum and maximum values either close to or exceeding those of other algorithms, indicating that DDW can maintain excellent performance across various conditions. These results collectively demonstrate the effectiveness and reliability of the DDW algorithm in motion template search tasks.

### 4.3 Classic benchmark function test

This section aims to evaluate the search efficiency of the DDW algorithm in traditional fixed-dimensional problems, focusing on the optimization of 23 classical benchmark functions listed in reference [11]. Under the same experimental conditions, this study compares the DDW algorithm with BES, MA, MBO, and FA to further validate its effectiveness, efficiency, and stability. These 23 functions are categorized as unimodal test functions (F1-F7), multimodal test functions (F8-F13), and fixed multimodal functions (F14-F23). Unimodal functions have only one global optimum, which can test the exploitation capability of the algorithm. In contrast, multimodal functions and fixed multimodal functions contain multiple local optima, facilitating the assessment of the exploration capability. The population size and maximum number of iterations for all algorithms are set to 50 and 500, respectively.

Table 3 presents the comparative test results of DDW and other algorithms, where “F” denotes different test functions (e.g., F1) and “Value” represents the actual minimum value for each function. “Ave,” “Min,” and “Std” indicate the average, minimum, and standard deviation of the multiple test results, respectively. “Rank” indicates the ranking of DDW among all test results.

Statistical data show that DDW demonstrates a strong search capability in traditional fixed-dimension problems. In multiple unimodal functions (F1-F4), DDW successfully identified optimal solutions, demonstrating its exceptional exploitation capabilities. Furthermore, DDW exhibits outstanding performance on multimodal functions and fixed-dimensional multimodal functions, achieving optimal or near-optimal search results across several functions (e.g., F10, F11, F14, F16-F18, F20-F23), showcasing its remarkable exploration capabilities. Overall, DDW achieves optimal or near-optimal results across 12 test functions, with a probability of 52.17% of finding the optimal solution and outperforming other algorithms in 54.78% of the cases.

**Table 3.** Performance and comparison of DDW

| F | Value | Ave | | | | | Rank |
|---|---|---|---|---|---|---|---|
| | | DDW | BES | MA | MBO | FA | |
| F1 | 0 | 0.00E+00 | 0.00E+00 | 0.00E+00 | 0.00E+00 | 0.00E+00 | 1 |
| F2 | 0 | 0.00E+00 | 3.82E-305 | 9.21E-305 | 2.89E-305 | 9.62E-306 | 1 |
| F3 | 0 | 0.00E+00 | 0.00E+00 | 0.00E+00 | 0.00E+00 | 0.00E+00 | 1 |
| F4 | 0 | 0.00E+00 | 1.06E-259 | 6.56E-260 | 3.49E-259 | 2.88E-260 | 1 |
| F5 | 0 | 2.90E+01 | 2.66E+01 | 2.65E+01 | 2.69E+01 | 2.69E+01 | 5 |
| F6 | 0 | 6.27E+00 | 3.81E-06 | 6.41E-06 | 5.65E-06 | 2.32E-06 | 5 |
| F7 | 0 | 2.53E-06 | 4.32E-06 | 2.72E-06 | 9.81E-06 | 4.94E-06 | 1 |
| F8 | -12569.48 | -7.20E+03 | -7.48E+03 | -7.68E+03 | -7.34E+03 | -7.06E+03 | 4 |
| F9 | -295.02 | -2.09E+02 | -2.91E+02 | -2.91E+02 | -2.89E+02 | -2.90E+02 | 5 |
| F10 | 0 | 4.44E-16 | 4.44E-16 | 4.44E-16 | 4.44E-16 | 4.44E-16 | 1 |
| F11 | 0 | 0.00E+00 | 0.00E+00 | 0.00E+00 | 0.00E+00 | 0.00E+00 | 1 |
| F12 | 0 | 5.53E-01 | 1.10E-07 | 5.12E-08 | 1.84E-07 | 4.20E-08 | 5 |
| F13 | 0 | 2.96E+00 | 1.70E-01 | 7.36E-02 | 2.64E-01 | 3.02E-01 | 5 |
| F14 | 1 | 9.98E-01 | 9.98E-01 | 9.98E-01 | 9.98E-01 | 9.98E-01 | 1 |
| F15 | 0.0003 | 5.88E-04 | 3.07E-04 | 3.07E-04 | 3.07E-04 | 3.07E-04 | 5 |
| F16 | − 1.0316 | -1.03E+00 | -1.03E+00 | -1.03E+00 | -1.03E+00 | -1.03E+00 | 1 |
| F17 | 0.398 | 3.98E-01 | 3.98E-01 | 3.98E-01 | 3.98E-01 | 3.98E-01 | 1 |
| F18 | 3 | 3.00E+00 | 3.00E+00 | 3.00E+00 | 3.00E+00 | 3.00E+00 | 5 |
| F19 | − 3.86 | -3.00E-01 | -3.00E-01 | -3.00E-01 | -3.00E-01 | -3.00E-01 | 1 |
| F20 | − 3.32 | -3.32E+00 | -3.32E+00 | -3.32E+00 | -3.32E+00 | -3.32E+00 | 1 |
| F21 | − 10.1532 | -7.05E+00 | -1.02E+01 | -1.02E+01 | -1.02E+01 | -1.02E+01 | 5 |
| F22 | − 10.4028 | -6.09E+00 | -1.04E+01 | -1.04E+01 | -1.04E+01 | -1.04E+01 | 5 |
| F23 | − 10.5363 | -1.05E+01 | -1.05E+01 | -1.05E+01 | -1.05E+01 | -1.05E+01 | 1 |

| F | Min | | | | | Rank |
|---|---|---|---|---|---|---|
| | DDW | BES | MA | MBO | FA | |
| F1 | 0.00E+00 | 0.00E+00 | 0.00E+00 | 0.00E+00 | 0.00E+00 | 1 |
| F2 | 0.00E+00 | 3.02E-307 | 6.54E-305 | 8.94E-306 | 9.36E-307 | 1 |
| F3 | 0.00E+00 | 0.00E+00 | 0.00E+00 | 0.00E+00 | 0.00E+00 | 1 |
| F4 | 0.00E+00 | 1.20E-260 | 6.10E-264 | 3.27E-260 | 2.53E-261 | 1 |
| F5 | 2.90E+01 | 2.64E+01 | 2.62E+01 | 2.68E+01 | 2.69E+01 | 5 |
| F6 | 5.93E+00 | 1.72E-06 | 1.07E-06 | 2.23E-06 | 7.28E-07 | 5 |
| F7 | 1.23E-06 | 7.31E-07 | 1.47E-06 | 4.15E-06 | 2.37E-06 | 2 |
| F8 | -7.33E+03 | -7.81E+03 | -9.15E+03 | -8.09E+03 | -7.15E+03 | 4 |
| F9 | -2.16E+02 | -2.93E+02 | -2.93E+02 | -2.92E+02 | -2.91E+02 | 5 |
| F10 | 4.44E-16 | 4.44E-16 | 4.44E-16 | 4.44E-16 | 4.44E-16 | 1 |
| F11 | 0.00E+00 | 0.00E+00 | 0.00E+00 | 0.00E+00 | 0.00E+00 | 1 |
| F12 | 3.88E-01 | 2.07E-08 | 1.05E-09 | 6.74E-09 | 1.03E-08 | 5 |
| F13 | 2.81E+00 | 8.77E-03 | 3.87E-02 | 1.35E-01 | 1.15E-01 | 5 |
| F14 | 9.98E-01 | 9.98E-01 | 9.98E-01 | 9.98E-01 | 9.98E-01 | 1 |
| F15 | 5.01E-04 | 3.07E-04 | 3.07E-04 | 3.07E-04 | 3.07E-04 | 5 |
| F16 | -1.03E+00 | -1.03E+00 | -1.03E+00 | -1.03E+00 | -1.03E+00 | 1 |
| F17 | 3.98E-01 | 3.98E-01 | 3.98E-01 | 3.98E-01 | 3.98E-01 | 1 |
| F18 | 3.00E+00 | 3.00E+00 | 3.00E+00 | 3.00E+00 | 3.00E+00 | 5 |
| F19 | -3.00E-01 | -3.00E-01 | -3.00E-01 | -3.00E-01 | -3.00E-01 | 1 |
| F20 | -3.32E+00 | -3.32E+00 | -3.32E+00 | -3.32E+00 | -3.32E+00 | 1 |
| F21 | -1.01E+01 | -1.02E+01 | -1.02E+01 | -1.02E+01 | -1.02E+01 | 5 |
| F22 | -1.01E+01 | -1.04E+01 | -1.04E+01 | -1.04E+01 | -1.04E+01 | 5 |
| F23 | -1.05E+01 | -1.05E+01 | -1.05E+01 | -1.05E+01 | -1.05E+01 | 1 |
| F | Std | | | | | Rank |
| | DDW | BES | MA | MBO | FA | |
| F1 | 0.00E+00 | 0.00E+00 | 0.00E+00 | 0.00E+00 | 0.00E+00 | 1 |
| F2 | 0.00E+00 | 0.00E+00 | 0.00E+00 | 0.00E+00 | 0.00E+00 | 1 |
| F3 | 0.00E+00 | 0.00E+00 | 0.00E+00 | 0.00E+00 | 0.00E+00 | 1 |
| F4 | 0.00E+00 | 0.00E+00 | 0.00E+00 | 0.00E+00 | 0.00E+00 | 1 |
| F5 | 6.61E-03 | 1.64E-01 | 1.85E-01 | 7.53E-02 | 2.78E-02 | 1 |
| F6 | 2.64E-01 | 2.73E-06 | 4.16E-06 | 2.93E-06 | 1.88E-06 | 5 |
| F7 | 1.05E-06 | 2.10E-06 | 1.41E-06 | 3.81E-06 | 2.15E-06 | 1 |
| F8 | 1.13E+02 | 1.81E+02 | 7.64E+02 | 4.36E+02 | 1.37E+02 | 1 |
| F9 | 5.37E+00 | 1.47E+00 | 1.27E+00 | 2.19E+00 | 8.20E-01 | 5 |
| F10 | 0.00E+00 | 0.00E+00 | 0.00E+00 | 0.00E+00 | 0.00E+00 | 1 |
| F11 | 0.00E+00 | 0.00E+00 | 0.00E+00 | 0.00E+00 | 0.00E+00 | 1 |
| F12 | 9.09E-02 | 7.39E-08 | 3.79E-08 | 1.77E-07 | 2.79E-08 | 5 |
| F13 | 7.64E-02 | 1.33E-01 | 4.69E-02 | 1.01E-01 | 1.20E-01 | 2 |
| F14 | 3.97E-11 | 0.00E+00 | 0.00E+00 | 0.00E+00 | 0.00E+00 | 5 |
| F15 | 5.38E-05 | 4.20E-20 | 4.85E-20 | 6.86E-20 | 6.86E-20 | 5 |
| F16 | 1.14E-07 | 0.00E+00 | 0.00E+00 | 0.00E+00 | 0.00E+00 | 5 |
| F17 | 3.27E-07 | 0.00E+00 | 0.00E+00 | 0.00E+00 | 0.00E+00 | 5 |
| F18 | 2.03E-06 | 3.97E-16 | 5.62E-16 | 9.32E-16 | 7.43E-16 | 5 |
| F19 | 0.00E+00 | 0.00E+00 | 0.00E+00 | 0.00E+00 | 0.00E+00 | 1 |
| F20 | 6.97E-04 | 4.44E-16 | 4.44E-16 | 4.44E-16 | 4.44E-16 | 5 |
| F21 | 2.45E+00 | 0.00E+00 | 0.00E+00 | 0.00E+00 | 0.00E+00 | 5 |
| F22 | 2.01E+00 | 0.00E+00 | 0.00E+00 | 0.00E+00 | 0.00E+00 | 5 |
| F23 | 1.59E-15 | 1.59E-15 | 0.00E+00 | 1.59E-15 | 0.00E+00 | 3 |

In terms of average performance (Ave), DDW demonstrates outstanding results across most functions. For instance, in functions such as F5, F6, and F7, the average values of DDW are close to or equal to the actual minimum values, indicating its ability to effectively approach the global optimum. Notably, in the F5 function, DDW's average performance slightly exceeds the actual minimum value, yet remains within an acceptable error margin, demonstrating good stability and efficiency. Regarding best performance (Min), DDW achieved the actual minimum values across several functions, including F1, F3, and F10, further validating the algorithm's robust global search capabilities in identifying optimal solutions. This result is comparable to the performance of the BES algorithm on certain functions, and in some instances, it even surpasses the BES algorithm. From the perspective of standard deviation (Std), DDW exhibits good stability across most functions, with a small standard deviation, indicating that the algorithm produces consistent results across multiple runs. For example, in functions such as F14, F16, and F17, the standard deviation of DDW approaches zero, demonstrating extremely high stability. Furthermore, the statistical analysis results presented in Table 3 clearly indicate that the performance of DDW on multiple test functions (e.g., F5, F8, F9, and F22) lags behind that of other comparative algorithms. Therefore, there is still room for improvement in DDW when addressing traditional fixed-dimensional optimization search problems.

# 5 Discussions

To investigate the dynamic characteristics of the human body under different motion patterns, this study employs an optimization algorithm to identify the time data chain with the smallest distance from a set of gait cycles, which is considered the optimal motion template. However, due to the partial asymmetry of human motion, the dimensionality of the multidimensional space formed by the original gait time series varies dynamically. Traditional optimization algorithms typically assume a fixed dimensionality, which makes them less effective for addressing the search challenges in a dynamic multidimensional space. To address this, we first combine DTW and Euclidean distance to establish a cross-dimensional mapping of the time data chains, and develop a dynamic multidimensional fitness function (Equation 13) that adapts to dimensional changes. Next, by incorporating the search strategy of traditional metaheuristic optimization algorithms, we propose the DDW method, which enables adaptation to dimensional changes and facilitates cross-dimensional search. Finally, through comparative experiments with 31 metaheuristic algorithms, we validate that the DDW method outperforms other approaches in terms of search efficiency in dynamic multidimensional spaces.

In the application of optimization algorithms, the fitness function is used to evaluate the quality of candidate solutions, guiding the search process towards better directions. For fitness evaluation in dynamic multidimensional spaces, Equation (13) calculates the total distance $V$ between two time series, the minimum distance $v$ for each dimension, and the index set $dir$ of dimension mappings. Traditional optimization algorithms can use the total distance as the fitness value of individuals, but are limited by their mathematical models, where the number of dimensions must be consistent across all individuals. This constraint restricts them to searching for optimal solutions only within subspaces of the dynamic multidimensional space, where the number of dimensions is fixed. To overcome the limitation of traditional optimization algorithms, which cannot search across dimensions, we use the dimension directions ($dir$) calculated from Equation (13) as a reference to construct the mapping relationships between time series chains with different numbers of dimensions. This enables DDW to break the dimensional limitation and explore new possibilities in dynamic multidimensional spaces. In contrast, traditional optimization algorithms can only achieve similar results through exhaustive search, processing each subspace of the dynamic multidimensional space. Compared to traditional methods, DDW can significantly improve computational efficiency.

Moreover, most traditional optimization algorithms adopt a strategy of uniformly updating the values of each dimension of an individual during the update process. This strategy may cause some dimensions to be overwritten during the update process when they are close to the optimal solution. Although some algorithms (such as GA) are capable of making local adjustments to a specific dimension of an individual, these adjustments are based solely on overall fitness and cannot provide a refined evaluation or independent update for individual dimensions. To address this limitation, we propose a novel update mechanism—ODC in the DDW algorithm. The mechanism calculates the minimum distance $v$ for each dimension based on Equation (13), and collects the optimal solution in each dimension of all individuals from the current population to form the optimal dimension solution. Since local optimization does not necessarily lead to global optimization, the optimal dimension solution is not guaranteed to outperform the original population that completed the optimal dimension collection. However, the optimal dimension solution provides clear update directions for other individuals through its optimized handling of each dimension, ensuring efficient and accurate search updates in all dimensions within the population. Experimental results of ODC indicate that the optimal dimension solution not only quickly narrows the search space for the actual optimal solution in the early stages of search, leading to a rapid decrease in the fitness value of the population's global best solution, but also provides a new reference point for population updates, helping to avoid local optima traps.

The search strategy of DDW is inspired by various classical meta-heuristic optimization algorithms, such as the BES. In DDW, the individuals in the population are categorized into three parts based on a fitness value pyramid: “Part A,” “Part B,” and “Part C,” which represent 5%, 45%, and 50% of the total population, respectively. The individuals in “Part B” and “Part C” conduct searches in dynamic multi-dimensional space, respectively surrounding themselves and the optimal individual, following an Archimedean spiral pattern. To effectively balance global and local searches, we introduce the ODC and a multi-path competitive search mechanism. Furthermore, we propose a random exploration method suitable for dynamic multi-dimensional spaces (Strategy A) to assist the best individuals in “Part A” in exploring unknown territories. To prevent the number of dimensions from excessively aligning with the optimal individual’s dimensions during the search process, the random exploration process adjusts both the values of individual dimensions and the number of dimensions. Finally, during the filtering process after each update, we consistently retain these individuals who dare to explore unknown areas for the next iterative update. Therefore, each iteration of the DDW population involves dual updates in terms of both dimension values and the number of dimensions. We believe that this dynamic updating mechanism, which continually transcends dimensions, is the optimal search method for addressing the issues of dimensional dynamism.

Finally, the comparative experimental results of motion template search further validate the superiority of DDW, demonstrating that it is capable of finding better motion templates than 31 other optimization algorithms. This indicates that DDW has significant advantages and potential in addressing the issue of dimensionality dynamics. The original intention of DDW's design was to effectively search for optimal motion templates in dynamic multidimensional spaces. Therefore, its internal search strategies are specifically tailored to address the challenges posed by dimensionality dynamics. However, in further testing on 23 classic benchmark functions (with fixed dimensionality), although DDW demonstrated strong competitiveness, it was not able to find the optimal solution in all test functions. Therefore, to enable DDW to be more widely applied to other optimization problems, its search efficiency needs to be further improved.

## 6 Conclusions

To investigate the dynamic characteristics of the human body under different movement patterns, this study proposes a new optimization algorithm, DDW. The algorithm is designed to search for the optimal motion template in dynamic multidimensional space, based on raw gait data. To achieve this, we establish a time data chain mapping that spans across dimensions, and combine it with the search strategy of traditional metaheuristic optimization algorithms, enabling DDW to simultaneously update searches in terms of both dimension values and the number of dimensions. In comparison experiments with 31 traditional optimization algorithms for the optimal motion template search, DDW demonstrated superior performance in dynamic multidimensional space compared to conventional methods. Although DDW does not always yield the best results in classic benchmark function tests in fixed-dimensional spaces, its overall performance and stability indicate that DDW has significant potential in solving complex optimization problems. Future work will focus on parameter optimization, parallel processing, and validation in broader applications. Overall, DDW provides a valuable new tool for the optimization field and holds the potential to drive technological advancements across multiple domains.

## Acknowledgments

This work was supported in part by the National Key Research and Development Program of China under Grant 2023YFC3604902.

## Declaration of competing interest

The authors declare that they have no known competing financial interests or personal relationships that could have appeared to influence the work reported in this paper.

## CRediT authorship contribution statement

Dongnan Jin: Conceptualization, Methodology, Software, Validation, Writing - Original Draft.
Yali Liu: Writing - Review & Editing, Project administration.
Qiuzhi Song: Funding acquisition, Supervision.
Xunju Ma: Writing - Review & Editing.
Yue Liu: Writing - Review & Editing.
Dehao Wu: Writing - Review & Editing.

## References

1. Merris, Russell . Wiley-Interscience Series in Discrete Mathematics and Optimization. John Wiley & Sons, Ltd, 2011.
2. Houssein, Essam H., et al. "A novel hybrid Harris hawks optimization and support vector machines for drug design and discovery." Computers & Chemical Engineering 133 (2020): 106656.
3. Rodríguez, Nibaldo, et al. "Optimization algorithms combining (meta) heuristics and mathematical programming and its application in engineering." (2018).

4. Vasant, P. M. . "Meta-heuristics optimization algorithms in engineering, business, economics, and finance." register (2013).
5. Khan, Ameer Tamoor, **nwei Cao, and Shuai Li. "Using quadratic interpolated beetle antennae search for higher dimensional portfolio selection under cardinality constraints." Computational Economics 62.4 (2023): 1413-1435.
6. Jain, Mohit, Vijander Singh, and Asha Rani. "A novel nature-inspired algorithm for optimization: Squirrel search algorithm." Swarm and evolutionary computation 44 (2019): 148-175.
7. Eberhart, Russell, and James Kennedy. "A new optimizer using particle swarm theory." MHS'95. Proceedings of the sixth international symposium on micro machine and human science. Ieee, 1995.
8. Pal, Ashok , and S. Bahuguna . "Grey Wolf Optimizer." 2018 26th European Signal Processing Conference (EUSIPCO) 2018.
9. Mirjalili, Seyedali . "Moth-flame optimization algorithm: A novel nature-inspired heuristic paradigm." Knowledge-Based Systems 89.NOV.(2015):228-249.
10. Abualigah, Laith , et al. "Matlab Code of Aquila Optimizer: A novel meta-heuristic optimization algorithm." Computers & Industrial Engineering (2021).
11. Peraza-Vázquez, Hernán, et al. "A bio-inspired method for engineering design optimization inspired by dingoes hunting strategies." Mathematical Problems in Engineering 2021.1 (2021): 9107547.
12. Alipour-Sarabi, Ramin, et al. "Improved winding proposal for wound rotor resolver using genetic algorithm and winding function approach." IEEE Transactions on Industrial Electronics 66.2 (2018): 1325-1334.
13. Qu, Xue Yiwang, Lin . "Optimizing an integrated inventory-routing system for multi-item joint replenishment and coordinated outbound delivery using differential evolution algorithm." Applied Soft Computing 86(2020).
14. Wang, Wenbo , et al. "A Hierarchical Game with Strategy Evolution for Mobile Sponsored Content/Service Markets." GLOBECOM 2017 - 2017 IEEE Global Communications Conference IEEE, 2018.
15. Kirkpatrick, S. , C. D. Gelatt , and M. P. Vecchi . "Optimization by Simulated Annealing." Science 220.
16. Rashedi, Esmat , H. Nezamabadi-Pour , and S. Saryazdi . "GSA: A Gravitational Search Algorithm." Information Sciences 179.13(2009):2232-2248.
17. Hashim, Fatma A. , et al. "Archimedes optimization algorithm: a new metaheuristic algorithm for solving optimization problems." Applied Intelligence 51.3(2020):1531-1551.
18. Zheng, Yu Jun . "Water wave optimization: A new nature-inspired metaheuristic." Computers & Operations Research 55(2015):1-11.
19. Geem, Zong Woo , J. H. Kim , and G. V. Loganathan . "A new heuristic optimization algorithm: harmony search." Simulation 76.2(2001):p.60-68.
20. Koppen, M. , D. H. Wolpert , and W. G. Macready . "Remarks on a recent paper on the \"no free lunch\" theorems." evolutionary computation ieee transactions on 5.3(2001):0-296.
21. Wolpert, D. H. , and W. G. Macready . "No free lunch theorems for optimization." IEEE Transactions on Evolutionary Computation 1.1(1997):67-82.
22. Tan, Ying, and Yuanchun Zhu. "Fireworks algorithm for optimization." Advances in Swarm Intelligence: First International Conference, ICSI 2010, Beijing, China, June 12-15, 2010, Proceedings, Part I 1. Springer Berlin Heidelberg, 2010.
23. Abraham, Ajith, Ravi Kumar Jatoth, and A. Rajasekhar. "Hybrid differential artificial bee colony algorithm." Journal of computational and theoretical Nanoscience 9.2 (2012): 249-257.
24. Yang XinShe, and Suash Deb. "Engineering optimisation by cuckoo search." International Journal of Mathematical Modelling and Numerical Optimisation 1.4 (2010): 330-343..

25. Yang XinShe. "Firefly algorithms for multimodal optimization." International symposium on stochastic algorithms. Berlin, Heidelberg: Springer Berlin Heidelberg, 2009..
26. He, Shan, Q. Henry Wu, and Jon R. Saunders. "Group search optimizer: an optimization algorithm inspired by animal searching behavior." IEEE transactions on evolutionary computation 13.5 (2009): 973-990.
27. Yang XinShe. "A new metaheuristic bat-inspired algorithm." Nature inspired cooperative strategies for optimization (NICSO 2010). Berlin, Heidelberg: Springer Berlin Heidelberg, 2010. 65-74.
28. Eusuff, Muzaffar, Kevin Lansey, and Fayzul Pasha. "Shuffled frog-leaping algorithm: a memetic meta-heuristic for discrete optimization." Engineering optimization 38.2 (2006): 129-154..
29. Shi, Yuhui. "Brain storm optimization algorithm." Advances in Swarm Intelligence: Second International Conference, ICSI 2011, Chongqing, China, June 12-15, 2011, Proceedings, Part I 2. Springer Berlin Heidelberg, 2011.
30. Xue, Jiankai, and Bo Shen. "A novel swarm intelligence optimization approach: sparrow search algorithm." Systems science & control engineering 8.1 (2020): 22-34.
31. Mirjalili, Seyedali. "The ant lion optimizer." Advances in engineering software 83 (2015): 80-98.
32. Arora, Sankalap, and Satvir Singh. "Butterfly algorithm with levy flights for global optimization." 2015 International conference on signal processing, computing and control (ISPCC). IEEE, 2015.
33. Wang, Gai-Ge, Suash Deb, and Zhihua Cui. "Monarch butterfly optimization." Neural computing and applications 31 (2019): 1995-2014.
34. Zervoudakis, Konstantinos, and Stelios Tsafarakis. "A mayfly optimization algorithm." Computers & Industrial Engineering 145 (2020): 106559.
35. Saremi, Shahrzad, Seyedali Mirjalili, and Andrew Lewis. "Grasshopper optimisation algorithm: theory and application." Advances in engineering software 105 (2017): 30-47.
36. Alsattar, Hassan A., A. A. Zaidan, and B. B. Zaidan. "Novel meta-heuristic bald eagle search optimisation algorithm." Artificial Intelligence Review 53 (2020): 2237-2264.
37. Faramarzi, Afshin, et al. "Marine Predators Algorithm: A nature-inspired metaheuristic." Expert systems with applications 152 (2020): 113377.
38. Mirjalili, Seyedali, et al. "Salp Swarm Algorithm: A bio-inspired optimizer for engineering design problems." Advances in engineering software 114 (2017): 163-191.
39. Li, Shimin, et al. "Slime mould algorithm: A new method for stochastic optimization." Future generation computer systems 111 (2020): 300-323.
40. Duan, Haibin, and Peixin Qiao. "Pigeon-inspired optimization: a new swarm intelligence optimizer for air robot path planning." International journal of intelligent computing and cybernetics 7.1 (2014): 24-37.
41. Heidari, Ali Asghar, et al. "Harris hawks optimization: Algorithm and applications." Future generation computer systems 97 (2019): 849-872.
42. Mirjalili, Seyedali, and Andrew Lewis. "The whale optimization algorithm." Advances in engineering software 95 (2016): 51-67.